\documentclass[10pt,conference]{IEEEtran}
\IEEEoverridecommandlockouts
\usepackage{amsmath,amssymb,amsfonts}
\usepackage{algorithmic}
\usepackage{graphicx}
\usepackage{textcomp}
\usepackage{xcolor}

\usepackage{url}
\usepackage{hyperref}
\usepackage{colortbl}
\usepackage{tabularx}
\usepackage{multirow}
\usepackage[numbers]{natbib}

\def\BibTeX{{\rm B\kern-.05em{\sc i\kern-.025em b}\kern-.08em
    T\kern-.1667em\lower.7ex\hbox{E}\kern-.125emX}}
\begin{document}

\title{Language Models for Novelty Detection in System Call Traces}

\author{\IEEEauthorblockN{Quentin Fournier}
	\IEEEauthorblockA{\textit{Polytechnique Montreal} \\
		Montreal, Quebec H3T 1J4 \\
	quentin.fournier@polymtl.ca}
	\and
	\IEEEauthorblockN{Daniel Aloise}
	\IEEEauthorblockA{\textit{Polytechnique Montreal} \\
		Montreal, Quebec H3T 1J4 \\
	daniel.aloise@polymtl.ca}
	\and
	\IEEEauthorblockN{Leandro R. Costa}
	\IEEEauthorblockA{\textit{Polytechnique Montreal} \\
		Montreal, Quebec H3T 1J4 \\
	leandro.costa@polymtl.ca}
}

\maketitle

\begin{abstract}
	Due to the complexity of modern computer systems, novel and unexpected behaviors frequently occur. Such deviations are either normal occurrences, such as software updates and new user activities, or abnormalities, such as misconfigurations, latency issues, intrusions, and software bugs. Regardless, novel behaviors are of great interest to developers, and there is a genuine need for efficient and effective methods to detect them. Nowadays, researchers consider system calls to be the most fine-grained and accurate source of information to investigate the behavior of computer systems. Accordingly, this paper introduces a novelty detection methodology that relies on a probability distribution over sequences of system calls, which can be seen as a language model. Language models estimate the likelihood of sequences, and since novelties deviate from previously observed behaviors by definition, they would be unlikely under the model. Following the success of neural networks for language models, three architectures are evaluated in this work: the widespread LSTM, the state-of-the-art Transformer, and the lower-complexity Longformer. However, large neural networks typically require an enormous amount of data to be trained effectively, and to the best of our knowledge, no massive modern datasets of kernel traces are publicly available. This paper addresses this limitation by introducing a new open-source dataset of kernel traces comprising over 2 million web requests with seven distinct behaviors. The proposed methodology requires minimal expert hand-crafting and achieves an F-score and AuROC greater than 95\% on most novelties while being data- and task-agnostic. The source code and trained models are publicly available on GitHub\footnote{\url{https://github.com/qfournier/syscall_novelty_detection}} while the datasets are available on Zenodo\footnote{\url{https://zenodo.org/record/7378420}}.
\end{abstract}


\section{Introduction}
\label{sec:introduction}

Even though computer systems are virtually deterministic, complex interactions between hardware and software often result in novel and unexpected behaviors. Novel behaviors are deviations from what has been previously observed and may be common behaviors such as component upgrades, software updates, new users, and rare queries, or anomalies such as misconfigurations, latency, intrusions, hardware failures, and bugs. This research focuses on detecting novelties rather than anomalies since it is a broader problem and since normal yet novel behaviors are often of interest to practitioners. Moreover, anomaly detection methods may be ineffective in detecting clever attacks designed to resemble legitimate users, which may still be detected as novel behaviors.

A non-intrusive and lightweight approach to recording the behavior of computer systems is to trace them. Tracing is the act of collecting low-level events generated whenever a specific instruction called tracepoint is encountered at runtime. This research considers events generated by the operating system called kernel events since they expose the behavior of the whole system~\citep{yaghmour2000measuring}. Furthermore, kernel events allow tracing virtually any Linux system without modifying the source code since tracepoints are already implemented in the Linux kernel. In particular, this paper focuses on a subset of the kernel events named \emph{system calls} or syscall. System calls correspond to requests from applications running in the userspace to the kernel in order to access resources such as memory, network, or other devices that would otherwise be inaccessible. In short, system calls are the only way for an application to communicate with the operating system. As mentioned by \citet{kim2016lstm}, many researchers consider system calls to be the most fine-grained and accurate source of information to analyze computer systems.

Due to the computational speed of modern computers, operating systems often execute hundreds of system calls every second, making the manual analysis of a collection of kernel traces with tools such as Trace Compass extremely time-consuming. As a result, practitioners and researchers often analyze traces automatically~\citep{giraldeau2016wait}. Since novel behaviors are unexpected and unknown by definition, their detection is difficult to specify in practice. Consequently, novel behaviors are typically detected with machine learning techniques since they learn to solve the task from examples~\citep{tandon2005learning, asmitha2014a, murtaza2012on}. Nonetheless, most machine learning algorithms benefit from or require carefully hand-crafted features~\citep{murtaza2013a,fournier2019automatic}. For the past decade, research has focused on neural networks~\citep{kim2016lstm,dymshits2017process,nedelkoski2019anomaly} since they automatically learn to extract meaningful features for the task, thereby reducing the need for an expert and improving the performance. Since traces are sequences of discrete values comprising a syntax and a semantic akin to natural languages~\citep{fournier2021on}, deep learning techniques from natural language processing (NLP) are particularly well suited for traces.

Natural language processing is the use of natural language by a computer and includes various tasks such as question answering, machine translation, summarization, sentiment analysis, and image captioning. A wide range of NLP applications rely on a probability distribution over sequences of tokens, often words or characters, called a language model (LM)~\cite{goodfellow2016deep}. One of the most popular approaches to learning a language model is the left-to-right LM, whose objective is to predict the conditional probability of each token knowing the previous ones. Formally, given a sequence of $N$ tokens $\boldsymbol{w}=\{w_1,  \dots, w_N\}$, the left-to-right language model computes for each token $w_i$ the conditional probability $P(w_i|w_{i-1},\dots,w_1)$. The chain rule of probability states that the joint probability of the entire sequence is the product of all the conditional probabilities:
\begin{equation}
	P(w_1,\dots,w_N) = \prod_{i=1}^N P(w_i|w_{i-1},\dots,w_1)
\end{equation}
Neural network language models typically minimize the cross-entropy loss, which is equivalent to maximizing the joint probability of the sequence. In other words, such language models maximize the likelihood of the known behaviors. By definition, novel behaviors deviate from what has been previously observed and therefore have a low likelihood under a language model trained on known behaviors.

Previous research studied log and trace language models for anomaly and novelty detection~\citep{kim2016lstm,brown2018recurrent,du2017deeplog,bogatinovski2020self}. Our approach improves over and diverges from these existing approaches with three significant contributions: (1) the quality and quantity of the data are drastically improved, (2) neural networks that are able to learn extremely long dependencies are investigated, and (3) the novelty detection methodology takes into account the sequence length. We next expand on these three contributions.

First, deep learning approaches are known to greatly depend on the quality and quantity of the data~\cite{goodfellow2016deep}. Nonetheless, the public datasets considered by current research, such as UNM~\citep{forrest1996a}, KDD98~\citep{lippmann2000evaluating}, and ADFA-LD~\citep{creech2013generation}, are small and obsolete, as explained by \citet{creech2013generation} and \citet{murtaza2013a}. Furthermore, these datasets lack the system call arguments, a valuable piece of information that has been shown to improve the performance of neural networks for language models~\citep{fournier2021on}. As a result, these datasets are inadequate for effectively training large neural networks and are not representative of modern systems. As a solution, this paper introduces a massive dataset of kernel traces that includes system call arguments and comprises more than 2 million web requests from seven realistic scenarios, including misconfigurations and latencies. Notably, our dataset enables training larger neural networks, such as the Transformer, that have become state-of-the-art in other fields. The dataset has been made public, as well as the data collection methodology and the scripts for reproducibility.

Second, current anomaly and novelty detection approaches rely on recurrent neural networks (RNNs), most often the Long Short-Term Memory~\citep{hochreiter1997long} (LSTM) network. However, recurrent networks are unable to efficiently model long-term dependencies due to their iterative nature. As explained by \citet{khandelwal2018sharp}, LSTM language models sharply distinguish recent positions but only vaguely remember the distant past. \citet{dai2019transformer} estimated that the relative effective context length (RECL) of LSTMs on natural language is between 200 and 400 tokens, which is consistent with \citet{khandelwal2018sharp} estimation. This inherent limitation of recurrent networks is analyzed in the case of kernel traces since they are typically much longer, comprising thousands of events. In particular, this paper investigates the state-of-the-art network for sequence processing called the Transformer~\citep{vaswani2017attention}, whose main advantage is the ability to model dependencies of arbitrary length. However, this flexibility comes at the expense of a quadratic complexity with respect to the sequence length. Consequently, a linear-complexity alternative called the Longformer~\citep{beltagy2020longformer} is also investigated.

Third, anomaly and novelty detections are typically performed with a top-$k$ on the individual conditional probabilities~\citep{nedelkoski2019anomaly,du2017deeplog,bogatinovski2020self} or a threshold on the joint probability of the sequence~\citep{kim2016lstm,brown2018recurrent}. However, conditional probabilities are only able to detect deviations of single events, also known as point outliers, while the joint probability does not take into account the sequence length. As a solution, our methodology leverages the \emph{perplexity}, a prevalent measure of how well a probability model predicts a sample~\citep{vaswani2017attention, devlin2019bert}.

The remainder of this paper is organized as follows. Section~\ref{sec:related_work} surveys the related works. Section~\ref{sec:novelty_detection} introduces the proposed novelty detection methodology. Section~\ref{sec:data_collection} presents the dataset collection methodology and analyzes the dataset. Section~\ref{sec:results} details the experiments and reports the results of the proposed approach on the collected dataset. Section~\ref{sec:threats_to_validity} acknowledges internal and external validity threats. Section~\ref{sec:discussion} discusses the strengths and limitations of the proposed methodologies as well as suggests interesting future research avenues. Finally, Section~\ref{sec:conclusion} concludes this paper.

\section{Related Work}
\label{sec:related_work}

Let us preface the related works by discussing the distinction between anomaly and novelty. As defined in Section~\ref{sec:introduction}, a novelty is any deviation from previously observed behaviors. Novelties include anomalies since they are typically unknown and unexpected, but not all novelties are anomalies. For instance, new users and rare queries are novel yet normal behaviors. The vast majority of the literature focuses on anomalies, and one of the most popular approaches is learning a ``normal'' behavior from the data and identifying any deviations from this behavior as abnormal~\citep{kim2016lstm,brown2018recurrent,du2017deeplog,bogatinovski2020self,guo2021logbert}. We argue that these approaches would be better framed as novelty detection methods as an additional mechanism would be necessary to determine whether the novel behaviors are normal or abnormal. For that reason, even though this paper focuses on novelty detection, most of the approaches discussed in this section were published under the anomaly detection paradigm.

This section surveys the fundamental aspects of the relevant related works: the trace representation, the machine learning model, and the anomaly or novelty detection scheme. 

\subsection{Trace Representation}

A trace usually comprises millions of low-level events, each containing multiple arguments, making them resource-intensive to handle. As a result, researchers have traded information for compactness in three ways: reducing the number of arguments, aggregating the events across time, and extracting higher-level features.

The first and foremost approach is to reduce the number of arguments. Current research often exclusively considers the event names and ignores the arguments, such as the process name and the return value. However, arguments are valuable data that allow the model to make more informed and, ultimately, more accurate predictions. Indeed, temporal information such as the response time~\citep{nedelkoski2019anomaly,nedelkoski2019anomaly2}, the timestamp~\citep{fournier2021on}, and the duration~\citep{du2017deeplog} has recently been considered with great success. Instead of reducing the number of arguments, \citet{ezeme2020a} compressed the values of the arguments by encoding the characters using ASCII values and considering the frequency distribution of these values for each argument. Nonetheless, contemporary research demonstrated the benefit of considering the actual values of multiple system call arguments for neural language models~\citep{fournier2021on}.

The second approach is to aggregate the events across time. The main example of this approach is called bag-of-words, also known as system call counts vector~\citep{dymshits2017process}, frequency counts of system call names~\citep{liu2005a} or bag of system calls~\citep{kang2005learning}. A bag-of-words is a representation that describes the number of occurrences of each token within a document. For instance, consider a vocabulary $\mathbb{V}=\{a, b, c\}$ and a sequence $\boldsymbol{w} = \{ a, c, c, c, a, c\}$. The bag-of-word representation of this sequence is $[2,0,4]$. The aggregation trades the ordering and fine-grained temporal information for a more compact representation. However, temporal information may be critical to detecting some novelties, such as latency.

The third approach is to extract higher-level features from the trace, such as states of kernel modules~\citep{murtaza2013a} or execution states~\citep{fournier2019automatic}. Although carefully hand-crafted higher-level features may deliver excellent performance, they discard the fine-grained information that makes traces so valuable. Moreover, they are time-consuming, error-prone, and potentially suboptimal since they must often be hand-crafted specifically for the task considered. 

\subsection{Machine Learning Model}

Due to the widespread use of computer systems and the importance of detecting anomalies and novel behaviors, a wide range of machine learning techniques have been explored, including rule-based algorithms~\citep{tandon2005learning, tandon2006on}, naive Bayes~\citep{murtaza2012on,asmitha2014a}, decision trees~\citep{murtaza2012on,song2018exad}, hidden Markov models~\citep{murtaza2012on,xu2013a}, and support vector machines (SVM)~\citep{murtaza2012on}. Given the great success of deep learning, researchers have recently shifted toward a family of neural networks called recurrent neural networks (RNNs)~\citep{rumelhart1986schemata}.

Recurrent neural networks iteratively process variable-size sequences by sharing the parameters at each position. They have been successfully applied to a wide range of applications, including speech recognition~\citep{graves2013speech,graves2014towards}, image captioning~\citep{kiros2014multimodal}, machine translation~\citep{sutskever2014sequence}, and anomaly detection~\citep{kim2016lstm,du2017deeplog}. RNNs have the advantage of iteratively storing information in their memory, also referred to as hidden representation, allowing information from prior input tokens to influence the current output. However, RNNs suffer from vanishing and exploding gradient issues~\citep{bengio1993the}. As a solution, \citet{hochreiter1997long} introduced the now widely popular long short-term memory (LSTM) network, which mitigates these shortcomings with paths through time. Alternatively, \citet{cho2014learning} introduced the gated recurrent unit (GRU), which behaves and performs similarly to while requiring fewer parameters.

The LSTM and GRU have been at the core of numerous anomaly and novelty detection approaches. For instance, \citet{kim2016lstm} detected host-based intrusions with an ensemble of LSTMs trained on sequences of system call names. \citet{dymshits2017process} identified changes in the behavior of processes with unidirectional and bidirectional LSTMs trained on sequences of bag-of-words. \citet{song2018exad} identified and explained anomalies with an LSTM trained on time-series data obtained from traces. \citet{nedelkoski2019anomaly} detected anomalies with a multimodal network made of the concatenation of the hidden representations of two LSTMs trained on textual logs and real-valued response time. \citet{lv2018intrusion} detected intrusions by extending system-call sequences with a GRU.

Recurrent neural networks, including LSTMs, suffer from memory compression~\citep{cheng2016long} and are unable to model very long-term dependencies. In particular, LSTM language models only vaguely remember the distant past~\citep{khandelwal2018sharp} and are unable to model dependencies that span more than 200 to 400 tokens~\citep{khandelwal2018sharp, dai2019transformer}. One solution is to augment the LSTM with an attention mechanism. \citet{ezeme2020a} and \citet{brown2018recurrent} detected anomalies with LSTMs augmented with inter-attention. However, their inherently sequential nature prevents parallelizing them.

\begin{figure}[!htb]
	\centering
	\includegraphics[width=0.9\linewidth]{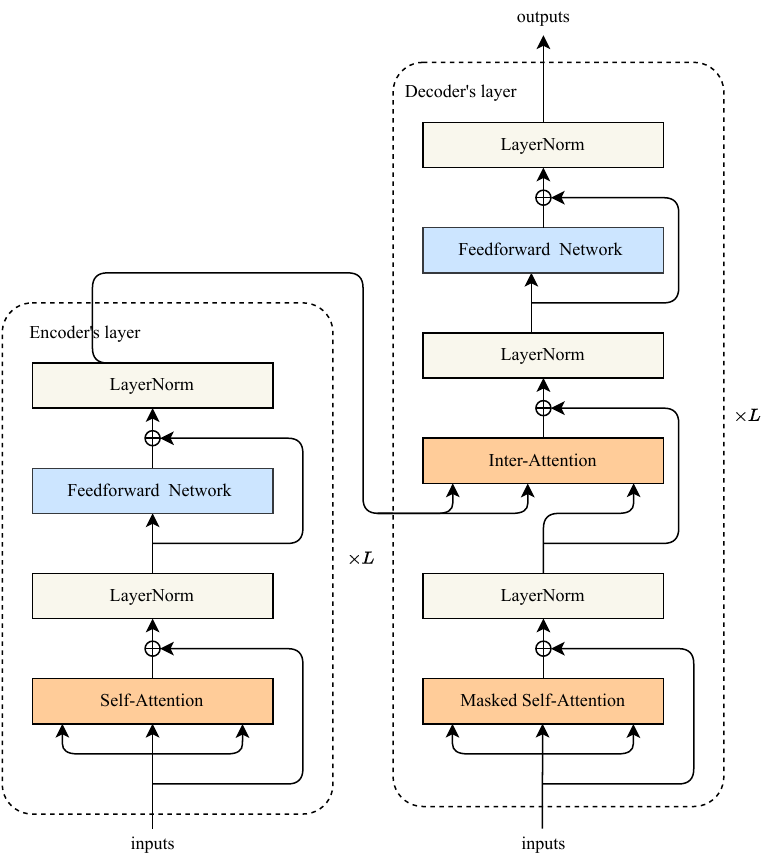}
	\caption{The computational graph of the Transformer, which comprises an encoder and a decoder. The encoder first processes the entire input sequence and produces a representation of each token attended by the decoder to generate the output sequence in an autoregressive manner.}
	\label{fig:Transformer}
\end{figure}

At the time of writing, LSTMs and GRUs have mostly been surpassed and replaced by the Transformer~\citep{vaswani2017attention}. As illustrated by Figure~\ref{fig:Transformer}, the Transformer is a simple network based solely on two attention mechanisms: the inter-attention and the self-attention. The latter computes a pairwise compatibility score between tokens corresponding to how much each token contributes to each output. As such, the self-attention replaces the role of recurrences in RNNs and enables modeling arbitrary length dependencies. Since this compatibility score is computed independently for each pair of tokens, the Transformer processes entire sequences simultaneously and can be efficiently parallelized. However, these major benefits come at the cost of quadratic complexity with respect to the sequence length. Consequently, a plethora of efficient alternatives have been proposed, such as the Longformer~\citep{beltagy2020longformer}, the Linformer~\citep{wang2020linformer}, and the Reformer~\citep{kitaev2020reformer}.
 
Despite the clear advantages and successes of the Transformer, researchers have yet to investigate this architecture to detect novelties in traces. For additional information on the Transformers, we refer the reader to the many surveys that have been published~\citep{galassi2020attention, weng2018attention, tay2020efficient, fournier2021a}.

\subsection{Detection Scheme}

Real-world datasets are often unlabeled regarding anomalies since they are typically unknown beforehand. Besides, manually labeling them afterward is generally time-consuming and error-prone. Consequently, the vast majority of approaches fall into the self-supervised or unsupervised setting. This paper focuses on language models such as the left-to-right LM that outputs the conditional probability of every possible token for each token in the sequence. In the literature, there are two distinct detection schemes based on the idea that novelties correspond to mispredictions.

The first scheme assumes that a misprediction occurs when the correct token does not appear in the top-$k$ most likely predictions. \citet{guo2021logbert} considered a sequence as anomalous if it contains more than a certain number of mispredictions, whereas \citet{bogatinovski2020self} considered the ratio of mispredictions. Temporal information is decisive in detecting some novelties, such as latencies. Consequently, in addition to the token mispredictions, \citet{du2017deeplog} and \citet{nedelkoski2019anomaly} predicted the timestamp and the response time, respectively.

The second scheme considers that a misprediction occurs when the conditional or joint probability is lower than a given threshold. Notably, \citet{kim2016lstm} and \citet{brown2018recurrent} detected anomalies with a threshold on the negative log-likelihood of the whole sequence.

\section{Methodology to Detect Novelties With Neural Language Models}
\label{sec:novelty_detection}

This section introduces the proposed methodology and follows the same structure as the literature review. 

\subsection{Trace Representation}

Neural networks learn to extract the relevant features for a task and thus typically benefit from richer inputs. Accordingly, this paper follows the methodology proposed in~\citep{fournier2021on} and relies on a joint representation of the system call name (\texttt{sysname}), the timestamp (\texttt{timestamp}), and five context fields that are added to all system calls by LTTng, namely the return value (\texttt{ret}), the process name (\texttt{procname}), the thread id (\texttt{tid}), the process id (\texttt{pid}), and whether the event corresponds to the start or the end of a system call execution (\texttt{entry}). 

To determine how to represent the arguments, one must first identify the inherently meaningful ones -- whose values convey meaning in themselves without any context. For explanatory purposes, let us consider a system call whose process name is \texttt{mysql} and whose process id is \texttt{15371}. The process name indicates that a MySQL database emitted the call, while the process id does not provide knowledge but allows relating the events emitted by the same process in the context of the trace. Out of the considered arguments, the \texttt{sysname}, \texttt{ret}, \texttt{entry}, and \texttt{procname} are inherently meaningful, while the \texttt{tid}, \texttt{pid}, and \texttt{timestamp} are not\footnote{There are exceptions, such as \texttt{pid} 0 and 1, which are meaningful.}.

The semantic knowledge contained in the values of the inherently meaningful arguments is encapsulated in a compact vectorial representation called embedding. An embedding effectively acts as a lookup table and is defined by a dense matrix $\boldsymbol{W} \in \mathbb{R}^{d_v \times d_e}$ where $d_v$ is the size of the vocabulary and $d_e$ is a hyperparameter corresponding to the dimension of the embedding such that $d_e \ll d_v$.

The values of the context-dependent arguments, such as the \texttt{pid} or \texttt{tid}, could be directly provided to the network as they are numerical values. However, it is best practice to normalize the input to mitigate potential numerical instabilities and speed up training~\citep{lecun2012efficient}. As a result, the context-dependent arguments are encoded with a succession of cosine and sine functions, as proposed by \citet{vaswani2017attention}. Formally, the encoding of a numerical value $x$ is a vector $\boldsymbol{pe}_x$ of dimension $d$ computed as $pe_{x,i} = \sin\left(x \times 10^{-6i/d}\right)$ if $i$ is even, otherwise $pe_{x,i} = \cos\left(x \times 10^{-6(i-1)/d)}\right)$, where $d$ is a hyperparameter.

In order to produce a joint representation of the system calls with their arguments and to provide a single input to the network, the embeddings and encodings must be combined. As mentioned in~\citep{fournier2021on}, the addition requires the vectors to have the same dimension and preserves that dimension, which may be too small to store all the information, thus creating a bottleneck. Consequently, the concatenation of the embeddings and encodings vectors is preferred.

Our methodology diverges from that of~\citep{fournier2021on} and~\citep{vaswani2017attention} in three aspects. First, the timestamps are converted into elapsed times between two consecutive system calls to avoid numerical instabilities as they exceed the largest value that can be stored on 32 bits. Second, the denominator of the encoding is increased from $10^4$ to $10^6$, as the values encoded are larger than in the work of \citet{vaswani2017attention}. Finally, the embeddings and encodings are all concatenated since this empirically resulted in more effective models.

\subsection{Neural Networks}

The proposed methodology was evaluated on a simple $n$-gram baseline, the widespread LSTM, the state-of-the-art Transformer, and the lower-complexity Longformer. Let us briefly introduce and justify each method.

The $n$-gram model makes the Markov assumption that the conditional probability may be approximated by only considering the $n-1$ tokens instead of all previous tokens. In other words, the $n$-gram model approximates the conditional probability $P(w_i|w_{i-1},\dots,w_1)$ as $P(w_i|w_{i-1}, \dots, w_{i-(n-1)})$, which is computed in practice as the number of times that\linebreak$\{w_{i-1}, \dots, w_{i-(n-1)}\}$ is followed by $w_i$ out of all the occurrences of $\{w_{i-1}, \dots, w_{i-(n-1)}\}$ in the dataset.

Following the vast majority of literature, the proposed methodology was evaluated using a unidirectional multi-layer LSTM. Since this architecture is well known and ubiquitous in the literature, the reader is referred to the original paper by \citet{hochreiter1997long} and the reference book of \citet{goodfellow2016deep} for a comprehensive description and analysis of the model.

As discussed in the introduction, kernel traces are typically much longer than the effective context length of LSTMs. As such, they may contain dependencies that the LSTM is unable to model. In order to investigate this potential limitation, the proposed methodology was evaluated on a vanilla Transformer.

The quadratic complexity of the Transformer means that the model cannot handle entire kernel traces in practice, even with multiple GPUs and mixed precision. Consequently, sequences were truncated, and the joint probability was estimated. In order to determine whether truncating sequences is a potential issue for kernel traces and to propose a solution that is more easily deployable in practice, a lower-complexity Transformer called the Longformer~\citep{beltagy2020longformer} was selected based on the attention patterns learned by the Transformer.

\begin{figure}[!htb]
	\centering
	\includegraphics[width=0.35\linewidth]{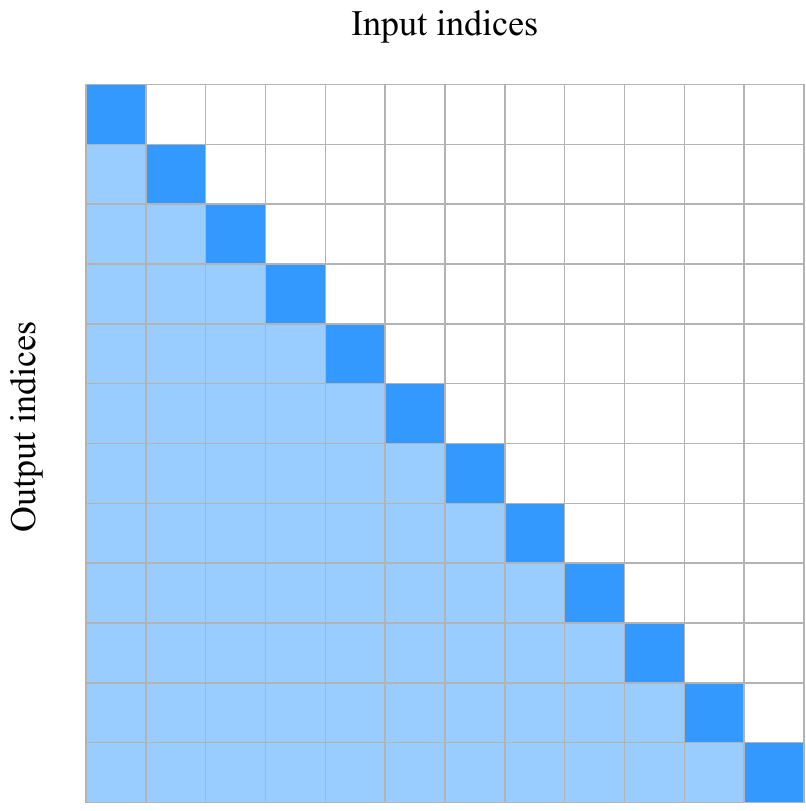}\hspace{2em}
	\includegraphics[width=0.35\linewidth]{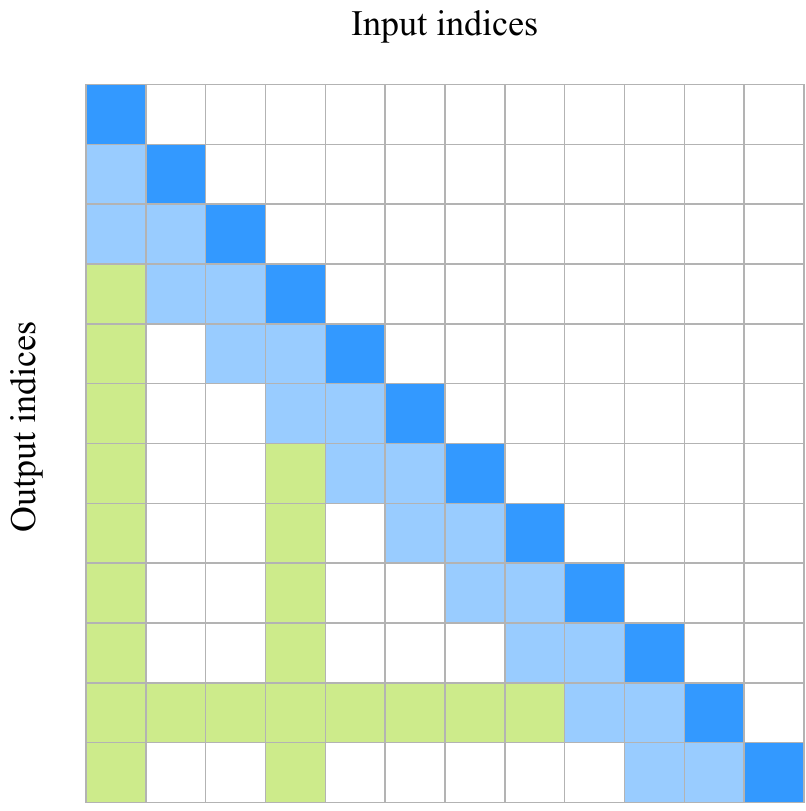}
	\caption{The connectivity matrices of the Transformer's full attention (left) and the Longformer's sparse attention (right). The window attention and global tokens are depicted in blue and green, respectively. The $i$-th output position attends to the $j$-th input position if, and only if, the cell $(i, j)$ is colored. The diagonal is highlighted to ease the reading. The matrices are lower triangular as future positions are masked to prevent the model from looking ahead at the solution.}
	\label{fig:attn}
\end{figure}

The Longformer achieves a linear complexity by replacing the self-attention mechanism with a combination of two sparse attention mechanisms called global tokens and sliding windows. Since the objective of the left-to-right language model is to predict the next token given the previous ones, future tokens are masked to prevent the model from looking ahead at the solution. Instead of looking at all previous tokens, the sliding window attention only considers the past $k$ tokens, similar to the $n$-gram model. Since only a fixed number of positions are considered for each token, the complexity of the window attention is linear with respect to the sequence length. Additionally, the global tokens are able to attend to every position and be attended by every position. Given that there is a fixed number of global tokens, each considering every token in the sequence, the complexity of global tokens is also linear with respect to the sequence length. Figure~\ref{fig:attn} depicts the full attention of the original Transformer and the sparse attention mechanisms of the Longformer. 

\subsection{Novelty Detection Scheme}
\label{sec:novelty_detection_scheme}

The neural networks are trained with the left-to-right language model, whose task is to predict the next token given the previous ones. Formally, given an input sequence $\boldsymbol{w}=\{w_1,\dots,w_N\}$ comprising $N$ tokens from a vocabulary $\mathbb{V}$, a neural left-to-right language model outputs the conditional probability $P(w^\ast|w_{i-1},\dots,w_1)$ for each $w^\ast$ in the vocabulary and for each position $i=1,\ldots,N$, such that $\sum_{w^\ast \in \mathbb{V}} P(w^{\ast}|w_{i-1},\dots,w_1) = 1$. The joint probability $P(w_1,\dots,w_N)$ is given by the chain rule of probability as $P(w_1,\dots,w_N) = \prod_{i=1}^N P(w_i|w_{i-1},\dots,w_1)$. 

However, since each conditional probability is lower than 1 in practice, longer sequences tend to have a lower joint probability. Let us consider an operating system that produces 200 system calls per second and two requests of 80 ms and 120 ms. Let us assume that the variation in response time is normal and that the events are all equally likely, with a conditional probability of 95\%. The two requests comprise $200 \times 0.08=16$ and $200 \times 0.12=24$ system calls, respectively. Consequently, their likelihood is $0.95^{16}=44\%$ and $0.95^{24}=29\%$, respectively. As a result, the likelihood of sequences is not well suited for novelty detection, as the throughput of system calls is high, and the sequence lengths may greatly vary. 

The \emph{perplexity} of a language model is a widely popular metric~\citep{vaswani2017attention,devlin2019bert} that measures its degree of uncertainty when a new token is generated, averaged over very long sequences. The per-word entropy $H$ of a sequence of word $\boldsymbol{w}=\{w_1,\dots,w_N\}$ generated by a language model is:
\begin{equation}
	H = \lim_{N\rightarrow \infty} -\frac{1}{N} \sum_{\boldsymbol{w}} P(\boldsymbol{w}) \log_2 P(\boldsymbol{w})
\end{equation}
Assuming ergodicity and given a large enough value of $N$, the summation may be discarded, and the entropy can be approximated as:
\begin{equation}
	\hat{H} = -\frac{1}{N} \log_2 P(\boldsymbol{w})
\end{equation}
Finally, the perplexity is defined as:
\begin{equation}
	PP =2^{\hat{H}} =P(\boldsymbol{w})^{-1/N} 
\end{equation}
where $N$ is the sequence length. In the above example, the perplexities of both requests are equal to $0.95^{-\frac{16}{16}}=0.95^{-\frac{24}{24}}=1.05$.

A sequence with a higher perplexity than the sequences in the training set is less likely under the model and can therefore be detected as a novelty. In practice, a simple threshold is efficient and provides excellent results. The threshold is empirically determined for each novel behavior with the in-distribution and out-of-distribution datasets to maximize the F-score.

\section{Data Collection}
\label{sec:data_collection}

Real-world traces are seldom released due to security and privacy concerns. Consequently, researchers often rely on the UNM~\citep{forrest1996a} and KDD98~\citep{lippmann2000evaluating} datasets. Nonetheless, these two datasets are over twenty years old and thus fail to represent modern systems~\citep{creech2013generation, murtaza2013a}. As a solution, \citet{creech2013generation} introduced ADFA-LD for host-based intrusion detection. However, ADFA-LD comprises only a few thousand samples without the system call arguments, which is too small for training large neural networks. Alternatively, \citet{murtaza2013a} introduced the much larger FirefoxDS dataset. Unfortunately, FirefoxDS is no longer available at the time of writing.

Neural networks greatly benefit from scaling as revealed by the current race toward ever-larger models~\citep{brown2020language,thoppilan2022lamda} and large networks greatly benefit from massive datasets~\citep{raffel2020exploring}. To the best of our knowledge, no modern and massive datasets of kernel traces are publicly available. This paper addresses this limitation by introducing a novel open-source dataset of kernel traces comprising over 2 million web requests with seven distinct behaviors. The dataset includes all the system calls arguments, and requests are well delimited by userspace events and labeled according to their behavior.

The remainder of this section explains the data collection methodology in detail and analyzes the collected dataset.

\subsection{Methodology}

\begin{figure}[!htb]
	\centering
	\includegraphics[width=\linewidth]{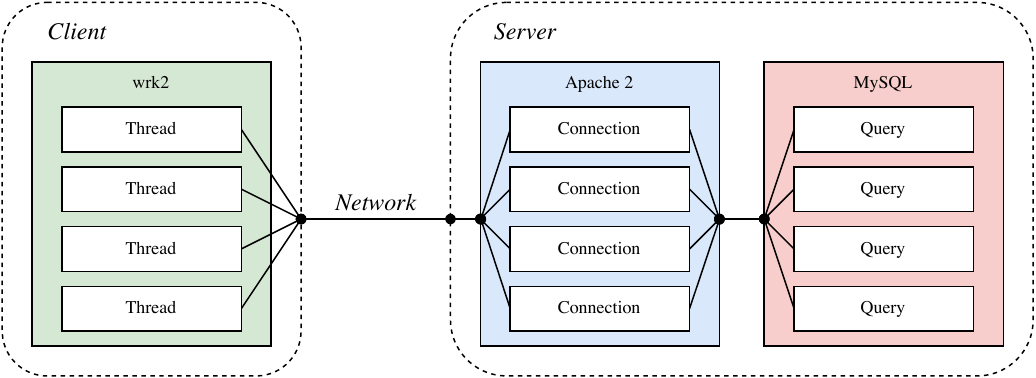}
	\caption{The client-server architecture.}
	\label{fig:client_server}
\end{figure}

Similar to the methodology of \citep{fournier2021on}, a benchmark tool sends numerous concurrent requests from the client to the server via the hypertext transfer protocol (HTTP). A web server receives the requests and calls PHP to query an SQL database and create the requested dynamic web page. The simple client-server architecture is depicted in Figure~\ref{fig:client_server}.

On the client side, the requests were emitted with the wrk2 benchmark tool, an open-source and multithreaded alternative to the Apache benchmark that guarantees a stable throughput for sufficiently long execution times. On the server side, the requests were handled with Apache2, a web server made popular thanks to its modular design. Apache2 was manually instrumented with two userspace events \texttt{httpd:enter\_event\_handler} and \texttt{httpd:exit\_event\_handler} to delimit requests. The requested dynamic web pages were created by querying MySQL with PHP installed as an Apache2 module. MySQL was chosen since it is an open-source relational database management system commonly used with Apache2. Finally, the database was filled with the Sakila sample database, as it is intended to provide a standard schema that can be used across numerous examples. Notably, this database comprises an author table with unique ids, first names, and last names.

As developers have limited access to the client side, this paper focuses on the server side, where most novelties originate. The kernel and userspace events were collected on the server with Linux Trace Toolkit: next generation (LTTng)~\citep{desnoyers2006the} due to its lightweight and rapidity~\citep{gebai2018survey}.

In order to identify novelties using a language model, the proposed methodology requires a set of known behaviors referred to as in-distribution (ID) and sets of novelties referred to as out-of-distribution (OOD). Accordingly, three ID sets were collected with a typical configuration under nominal load (train ID, validation ID, and test ID), and two OOD sets were collected for each of the following server-side novelties (validation OOD and test OOD).

\paragraph{CPU} The CPU is overloaded using the stress-ng tool, which performs numerous matrix multiplications. This behavior simulates a compute-intensive process competing for resources with the web server, which may arise from a cryptocurrency-mining procedure deployed by an intruder.

\paragraph{OPcache} The server is misconfigured by disabling PHP's OPcache, which stores precompiled script bytecode in memory to speed up the response time. This behavior may arise from a developer disabling the cache during the development phase and forgetting to re-enable it afterward.

\paragraph{Dump IO} The disk is overloaded by enabling the highest level of Apache2 log and storing them into a file using the dump\_io mod, which helps investigate the server's behavior but heavily uses storage resources and leads to a slower response time. Like OPcache, this behavior may arise from a developer enabling logging during development but forgetting to disable it afterward.

\paragraph{Connection} Apache2 is configured to support 150 concurrent connections by default and will start dropping requests as the traffic increases if that number is not increased. Instead of increasing the traffic, this behavior was reproduced by reducing the number of concurrent connections to 25, as the server would be IO-bounded before requiring more connections. Additionally, this approach allows for keeping the traffic consistent with the other behaviors.

\paragraph{Socket} The server is misconfigured by disabling Apache2 KeepAlive, which allows the web server and browsers to reuse the same socket for transferring multiple files, thereby reducing CPU usage at the cost of higher memory usage. By default, KeepAlive is enabled as CPU usage is often the main limiting factor. This behavior may arise when the server is redeployed from a memory-limited machine to a CPU-limited one.

\paragraph{SSL} The server's security is compromised by disabling the secure sockets layer (SSL), a protocol for establishing secure connections between web servers and browsers. This behavior may arise from a malicious intruder.

\subsection{Dataset Analysis}
\label{sec:dataset_analysis}

The web server was deployed on an Ubuntu 21.04 machine with 16-core Intel E5640 (up to 2.67 GHz) and 192 GB of RAM, and the client sent 1000 requests per second to load the server properly. The web server was traced for 1,000s for the training set and 100s for all validation and test sets. The dataset is balanced, although positive samples are expected to be much rarer than negative ones in real-world applications. This decision is justified in the threat to validity section.

In this work, requests comprise all the system calls generated between their start and end as delimited by the userspace events \texttt{httpd:enter\_event\_handler} and \texttt{httpd:exit\_event\_handler} regardless of their thread ids. Consequently, system calls may be added to multiple requests since they are concurrent. The primary reason behind this decision is that we want requests to include the events associated with the root cause of the novelties. For instance, let us assume that an unexpected process is taking CPU time, thus creating latency. To detect the root cause of this behavior, the events generated by this abnormal process must be included in the request, even though they do not have the same thread id. The main drawback of this approach is that requests contain significantly more events, making them more resource-intensive to process and increasing the noise. 

Table~\ref{tab:dataset} reports request statistics for each set after discarding the first second, corresponding to the initialization of LTTng. The lower number of requests for the CPU behavior is due to the server's inability to maintain the throughput due to the lack of CPU. Distributions of system call names, process names, length, and duration for each set are available on GitHub.

\begin{table*}[!htb]
	\caption{Statistics on the requests in each dataset.}
	\label{tab:dataset}
	\centering
	\begin{tabular}{llrrrrrrr}
		\hline
		\textbf{Behavior}           & \textbf{Dataset} & \textbf{Number of Requests} & \multicolumn{3}{c}{\textbf{Request Length}} & \multicolumn{3}{c}{\textbf{Request Duration} (ms)} \\
		                            &            &         & \multicolumn{1}{c}{min} & \multicolumn{1}{c}{mean} & \multicolumn{1}{c}{max} & \multicolumn{1}{c}{min} & \multicolumn{1}{c}{mean} & \multicolumn{1}{c}{max} \\
		\hline
		\multirow{3}{*}{ID}         & Train      & 999,063 & 238                     & 1105.7 ± 244.8          & 4,645                   & 0.28                    & 1.68 ± 0.65             & 53.61                   \\
		                            & Validation & 99,058  & 30                      & 1107.3 ± 244.9          & 2,803                   & 0.03                    & 1.67 ± 0.59             & 11.69                   \\
		                            & Test       & 99,065  & 240                     & 1108.7 ± 247.1          & 2,683                   & 0.91                    & 1.67 ± 0.61             & 12.36                   \\ \hline
		\multirow{2}{*}{Connection} & Validation & 99,016  & 246                     & 1125.7 ± 243.0          & 2,882                   & 0.94                    & 1.66 ± 0.60             & 15.02                   \\
		                            & Test       & 99,019  & 158                     & 1125.0 ± 243.3          & 2,792                   & 0.27                    & 1.66 ± 0.60             & 11.83                   \\ \hline
		\multirow{2}{*}{CPU}        & Validation & 57,616  & 258                     & 1910.6 ± 607.6          & 6,221                   & 1.31                    & 13.25 ± 6.06            & 52.10                   \\
		                            & Test       & 56,191  & 222                     & 1913.8 ± 596.0          & 6,363                   & 0.51                    & 13.58 ± 5.81            & 35.69                   \\ \hline
		\multirow{2}{*}{IO}         & Validation & 98,974  & 350                     & 1827.7 ± 323.4          & 6,155                   & 1.27                    & 2.13 ± 3.23             & 349.33                  \\
		                            & Test       & 98,980  & 392                     & 1821.1 ± 321.0          & 6,967                   & 1.25                    & 2.10 ± 1.23             & 103.69                  \\ \hline
		\multirow{2}{*}{OPcache}    & Validation & 99,069  & 256                     & 1162.9 ± 244.2          & 2,824                   & 0.99                    & 1.78 ± 0.60             & 14.79                   \\
		                            & Test       & 99,057  & 250                     & 1160.6 ± 245.9          & 2,896                   & 0.96                    & 1.77 ± 0.60             & 11.94                   \\ \hline
		\multirow{2}{*}{Socket}     & Validation & 99,074  & 216                     & 2082.0 ± 362.5          & 8,463                   & 0.83                    & 6.89 ± 0.73             & 48.79                   \\
		                            & Test       & 99,084  & 679                     & 2081.8 ± 355.7          & 7,032                   & 3.63                    & 6.89 ± 0.64             & 19.61                   \\ \hline
		\multirow{2}{*}{SSL}        & Validation & 99,072  & 16                      & 1058.1 ± 229.1          & 3,230                   & 0.04                    & 1.48 ± 0.36             & 15.92                   \\
		                            & Test       & 99,067  & 238                     & 1054.8 ± 230.2          & 3,855                   & 0.80                    & 1.47 ± 0.38             & 22.23                   \\
		\hline
						            
	\end{tabular}
\end{table*}

\section{Results}
\label{sec:results}

\subsection{Language Models}

The language models were trained on a server with 2 x Intel Gold 6148 Skylake @ 2.4 GHz, 4 x Nvidia V100SXM2 16G, and 64GB of RAM.

Neural networks were trained with mixed precision due to the simplicity of implementing NVIDIA's Automatic Mixed-Precision, which accelerates training and reduces memory consumption by storing and computing the weights, activations, and gradients in half-precision~\citep{micikevicius2018mixed}. Furthermore, due to the quadratic complexity of the Transformer, they were trained with gradient checkpointing, which trades memory for computation by recomputing the activations during the backward pass instead of storing them in memory during the forward pass~\citep{chen2016training}. Additionally, the Transformer truncated the few sequences longer than 2048 system calls to avoid exceeding the GPUs memory.

The following hyperparameters were manually tuned in a greedy fashion for all networks: the depth and width of the models, the embedding size, the optimizer, the warmup steps, the label smoothing weight, the dropout probability, the number of updates without improvements before reducing the learning rate, and the number of updates before early stopping. Additionally, the number of heads, the SwiGLU activation function~\citep{shazeer2020glu}, and the T-fixup initialization~\citep{huang2020improving} were considered for the Transformer and the Longformer. Furthermore, the window size, the dilation, and the number of global tokens were also considered for the latter. Overall, over 80 distinct network configurations were evaluated, each requiring about a day of computation on the aforementioned server. The exhaustive list of hyperparameters for each model is available on GitHub.

Table~\ref{tab:training} reports the average cross-entropy and top-1 accuracy of the three networks on all sets. Note that the top-1 accuracy is defined as the proportion of accurately predicted tokens, that is, the number of times the most probable system call name corresponds to the actual one over the total number of predictions. Each experiment was reproduced five times with different seeds to mitigate the stochasticity. The cross-entropy on the ID sets is consistently and significantly lower than on the OOD sets, indicating that the networks have a higher degree of uncertainty when modeling the novel behaviors. The LSTM outperforms the two attention-based networks in terms of cross-entropy and accuracy, although they have the advantage of learning arbitrary length dependencies. As expected, increasing the width and depth of the two attention-based models improved their performance. For instance, increasing the depth from 2 to 6 layers and the width from 672 to 896 allowed reducing the cross-entropy of the Transformer from 0.907 to 0.719 on the ID test set, outperforming the LSTM. However, although larger models were better at generalizing due to their higher flexibility, they performed poorly on our downstream novelty detection task since they assigned a high likelihood to all behaviors. Since our goal is to detect novelties, only the smaller models are reported in Table~\ref{tab:training} and thereafter.

\begin{table*}[!htb]
	\caption{Training Performance of the Neural Networks.}
	\centering
	\label{tab:training}
	\begin{tabular}{lrrrrrr}
		\hline
		& \multicolumn{2}{c}{\textbf{LSTM}} & \multicolumn{2}{c}{\textbf{Transformer}} & \multicolumn{2}{c}{\textbf{Longformer}} \\
						                        
		\multicolumn{1}{c}{Dataset} & \multicolumn{1}{c}{Cross-Entropy} & \multicolumn{1}{c}{Accuracy} & \multicolumn{1}{c}{Cross-Entropy} & \multicolumn{1}{c}{Accuracy} & \multicolumn{1}{c}{Cross-Entropy} & \multicolumn{1}{c}{Accuracy} \\
		\hline
		Train                       & \textbf{0.714 ± 0.002}           & \textbf{0.764 ± 0.000}      & 0.891 ± 0.039                    & 0.701 ± 0.013               & 0.875 ± 0.008                    & 0.712 ± 0.003               \\
		Test ID                     & \textbf{0.720 ± 0.002}           & \textbf{0.762 ± 0.000}      & 0.907 ± 0.038                    & 0.696 ± 0.012               & 0.885 ± 0.010                    & 0.708 ± 0.004               \\
		Test Connection             & \textbf{0.812 ± 0.017}           & \textbf{0.737 ± 0.006}      & 1.274 ± 0.103                    & 0.605 ± 0.025               & 1.105 ± 0.018                    & 0.651 ± 0.006               \\
		Test CPU                    & 0.961 ± 0.027                    & 0.736 ± 0.010               & 1.155 ± 0.056                    & 0.685 ± 0.012               & \textbf{ 0.940 ± 0.022}          & \textbf{0.744 ± 0.005}      \\
		Test IO                     & 2.287 ± 0.185                    & 0.366 ± 0.037               & 2.993 ± 0.307                    & 0.232 ± 0.042               & \textbf{2.082 ± 0.150}           & \textbf{0.391 ± 0.036}      \\
		Test OPcache                & \textbf{1.127 ± 0.019}           & \textbf{0.669 ± 0.005}      & 1.302 ± 0.052                    & 0.607 ± 0.013               & 1.254 ± 0.024                    & 0.630 ± 0.007               \\
		Test Socket                 & \textbf{1.008 ± 0.033}           & \textbf{0.699 ± 0.007}      & 1.573 ± 0.138                    & 0.549 ± 0.029               & 1.223 ± 0.033                    & 0.636 ± 0.012               \\
		Test SSL                    & \textbf{0.906 ± 0.018}           & \textbf{0.716 ± 0.007}      & 1.495 ± 0.105                    & 0.550 ± 0.030               & 1.245 ± 0.027                    & 0.619 ± 0.010               \\
		\hline
	\end{tabular}
\end{table*}

\begin{figure}[!htb]
	\centering
	\includegraphics[width=0.5\linewidth]{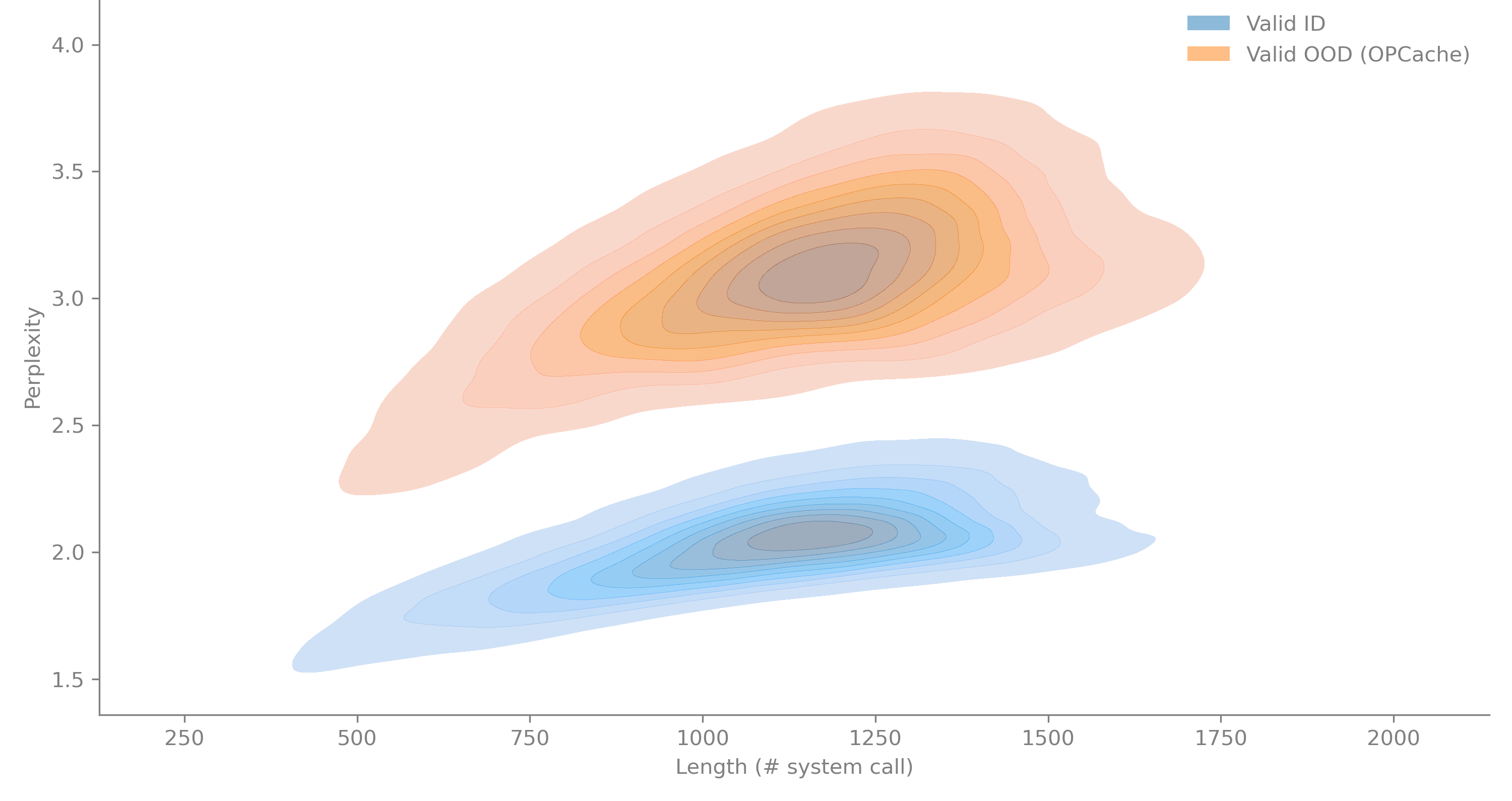}\hfill\includegraphics[width=0.5\linewidth]{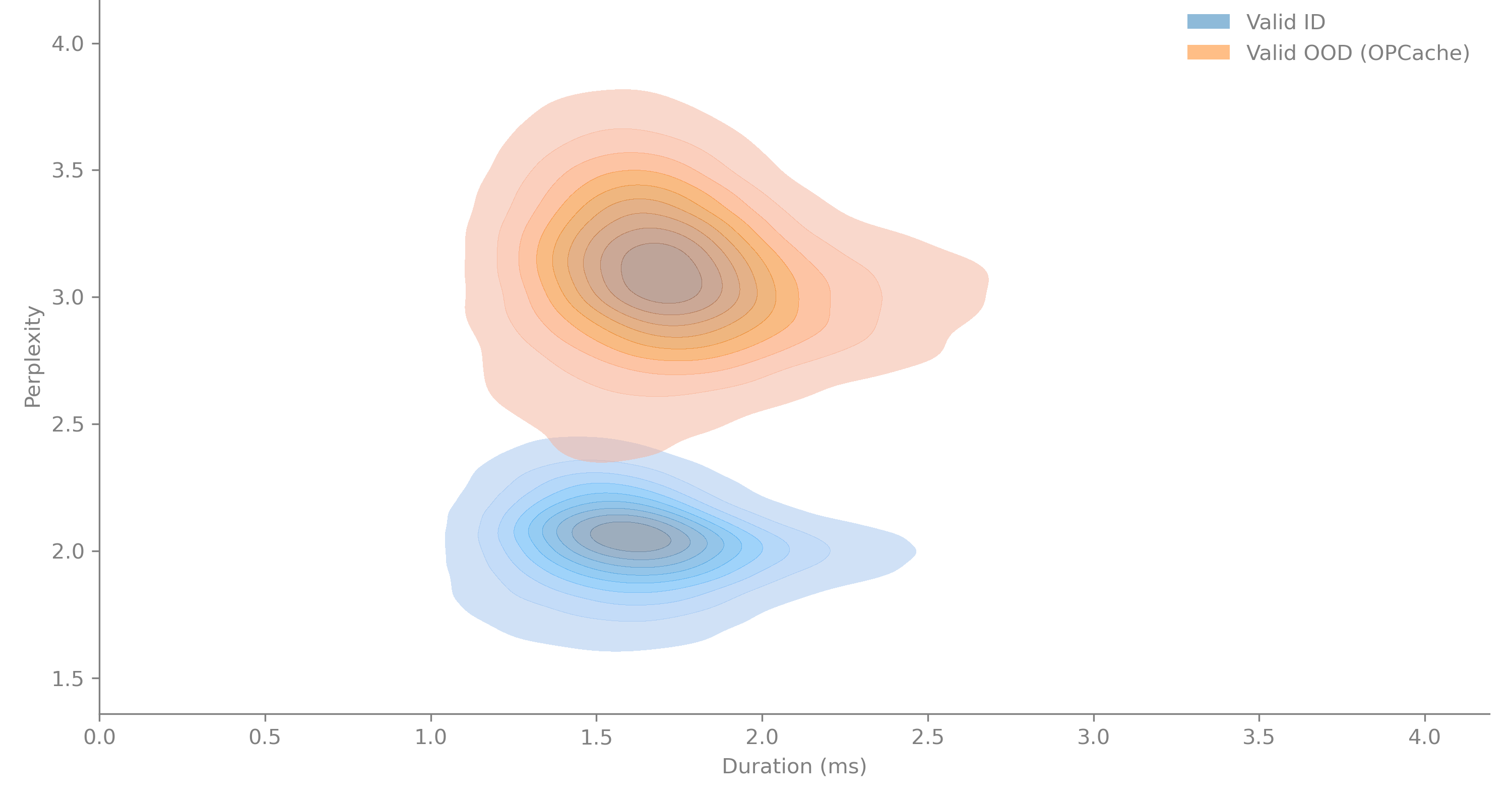}
	\caption{Distribution of the perplexity of requests in the ID (blue) and OPcache (orange) validation sets with respect to the length (left) and duration (right).} 
	\label{fig:perpexlity_distribution}
\end{figure}

Figure~\ref{fig:perpexlity_distribution} shows the perplexity distribution of requests in the ID and OPcache validation sets computed with the LSTM. Although the length and duration distributions are similar, the network assigns a higher perplexity to OOD requests, indicating the ability to leverage complex interactions between system calls. Similar figures for all datasets and models are available on GitHub.

\subsection{Novelty Detection}

Novelty detection is a binary classification problem since requests are either in-distribution or out-of-distribution. By convention, the outcomes of binary classification problems are referred to as positive and negative, with positive indicating the class of interest (i.e., novelties). Such problems are evaluated in terms of precision and recall, where the precision is the proportion of actual positives among the predicted positives, and the recall is the proportion of actual positives correctly predicted. Due to space constraints, the harmonic mean of the precision and recall, denoted F-score or F-measure, is instead reported. The classification is performed with a simple threshold on the perplexity, which acts as a novelty score, as described in Section| \ref{sec:novelty_detection_scheme}. The threshold that maximizes the F-score is first empirically determined for each validation set, and the F-score is then computed for each test set with the corresponding threshold. Additionally, the area under the ROC curve (AuROC) is reported. The ROC curve evaluates the ratio of true positives against the ratio of false positives at different thresholds instead of selecting a threshold to optimize a specific metric. For more information on the ROC curve, please refer to \citet{zou2007receiver}.

Table~\ref{tab:ood_detection} reports the AuROC and F-score of the 4-gram baseline and the three neural networks. The simple baseline could not detect the novel behaviors accurately, with the surprising exception of the IO behavior. The simplicity of detecting this behavior arises from the out-of-distribution requests having a wildly different distribution of system call names compared to the in-distribution request. Indeed, the two most common system calls in the training set are \texttt{recvfrom} (15\%) and \texttt{read} (10\%), while they are \texttt{write} (17\%) and \texttt{getpid} (17\%) for the IO dataset.

\begin{table*}[!htb]
	\caption{Novelty Detection Performance of the Language Models.}
	\centering
	\label{tab:ood_detection}
	\begin{tabular}{lrrrrrrrr}
		\hline
		& \multicolumn{2}{c}{\textbf{4-gram}} & \multicolumn{2}{c}{\textbf{LSTM}} & \multicolumn{2}{c}{\textbf{Transformer}} & \multicolumn{2}{c}{\textbf{Longformer}} \\
		Dataset         & \multicolumn{1}{c}{AuROC} & \multicolumn{1}{c}{F-score} & \multicolumn{1}{c}{AuROC} & \multicolumn{1}{c}{F-score} & \multicolumn{1}{c}{AuROC} & \multicolumn{1}{c}{F-score} & \multicolumn{1}{c}{AuROC} & \multicolumn{1}{c}{F-score} \\
		\hline
		Test Connection & 51.5                      & 66.7                        & 79.9 ± 3.8               & 74.8 ± 2.9                 & \textbf{97.8 ± 1.6}      & \textbf{94.3 ± 2.9}        & 93.6 ± 1.4               & 87.6 ± 1.8                 \\
		Test CPU        & 0.9                       & 53.2                        & \textbf{98.5 ± 0.6}      & \textbf{93.6 ± 2.1}        & 94.8 ± 1.8               & 85.1 ± 3.4                 & 67.9 ± 7.7               & 59.6 ± 4.1                 \\
		Test IO)        & 98.6                      & 94.7                        & \textbf{100.0 ± 0.0}     & \textbf{100.0 ± 0.0}       & \textbf{100.0 ± 0.0}     & \textbf{100.0 ± 0.0}       & \textbf{100.0 ± 0.0}     & \textbf{100.0 ± 0.0}       \\
		Test OPcache    & 65.2                      & 67.5                        & \textbf{99.7 ± 0.1}      & \textbf{98.3 ± 0.2}        & 99.1 ± 0.2               & 96.7 ± 0.4                 & 98.9 ± 0.2               & 96.2 ± 0.4                 \\
		Test Socket     & 22.6                      & 66.7                        & 98.8 ± 0.6               & 94.7 ± 2.3                 & \textbf{99.9 ± 0.1}      & \textbf{99.1 ± 0.6}        & 99.1 ± 0.4               & 96.4 ± 1.3                 \\
		Test SSL        & 50.5                      & 66.7                        & 91.9 ± 1.7               & 85.4 ± 2.0                 & \textbf{99.7 ± 0.2}      & \textbf{98.5 ± 0.9}        & 98.4 ± 0.5               & 94.5 ± 0.9                 \\
		\hline
	\end{tabular}
\end{table*}

\begin{figure}[!htb]
	\centering
	\includegraphics[width=0.35\linewidth]{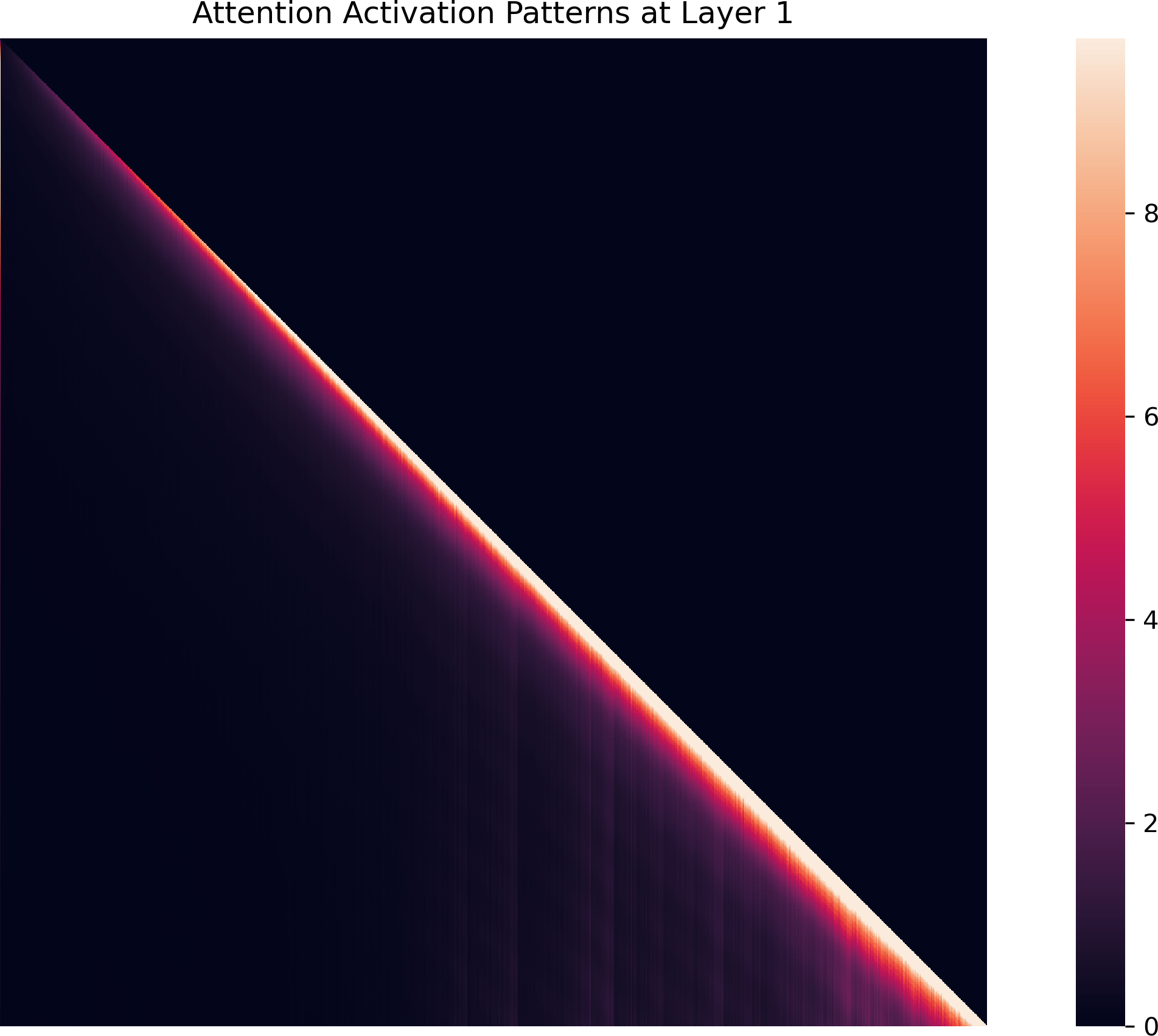}\hspace{2em}
	\includegraphics[width=0.35\linewidth]{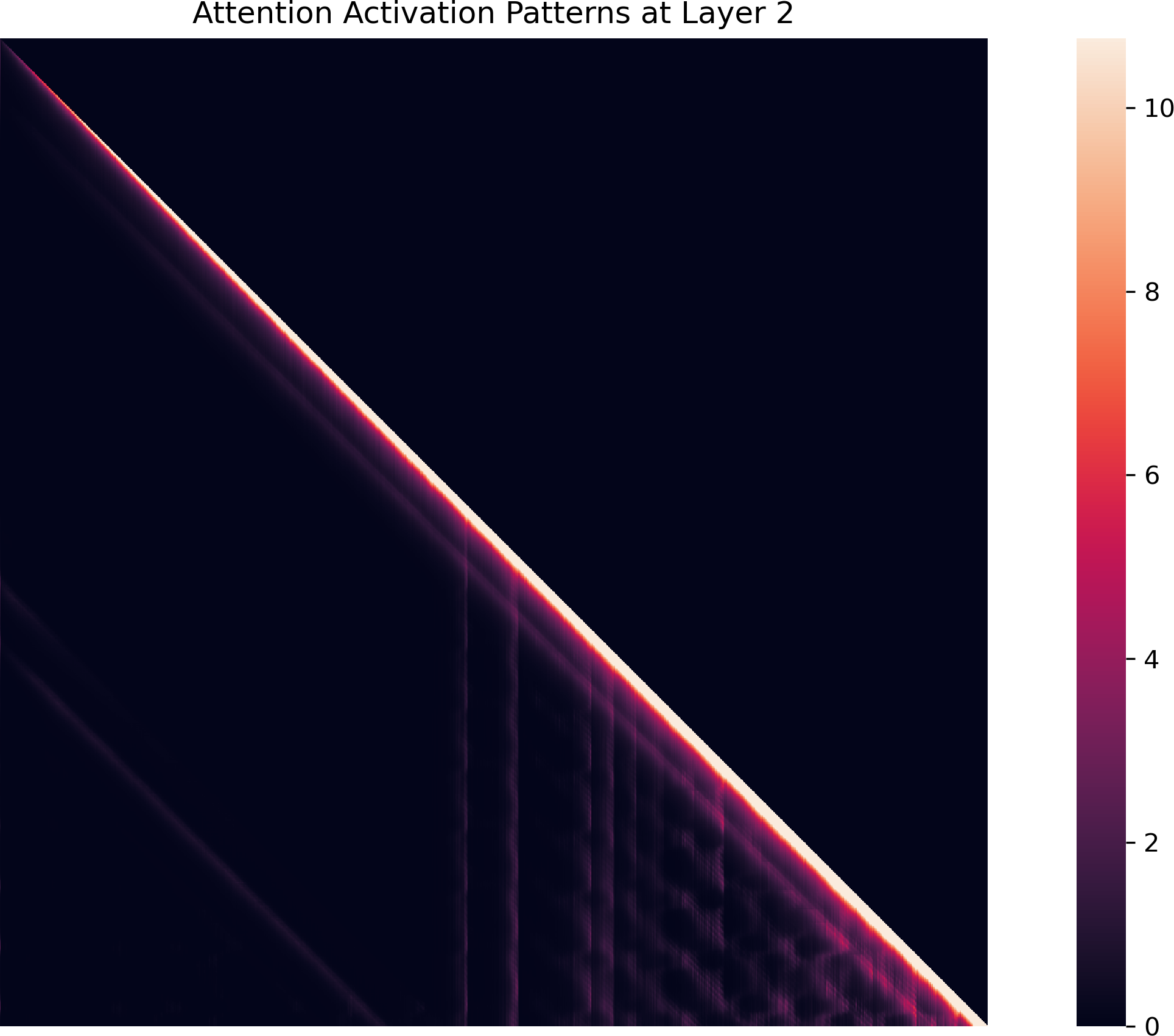}
	\caption{Attention activation patterns of the two layers of the Transformer on the ID validation set. For the sake of readability, the attention activation patterns are shown for the first 1024 positions and are compensated by multiplying each row with the number of unmasked positions.}
	\label{fig:attention_activation_patterns}
\end{figure}

Due to the recent successes of the Transformer over the LSTM, one would have expected the former to outperform the latter. However, that is not the case: the LSTM performed on par or better than the attention-based networks in 3 out of 6 behaviors. Figure~\ref{fig:attention_activation_patterns} illustrates the attention activation patterns of the Transformer on the in-distribution validation set. Interestingly, the dependencies modeled by the Transformer are mostly local since most of the attention learned is along the diagonal. This observation justifies the choice of the less complex Longformer, which relies on the window attention mechanism, thus assuming local dependencies. Furthermore, this observation explains the performance of the LSTM since RNNs are inherently biased toward local dependencies due to their iterative nature, making them well-suited for this use case. Nonetheless, in cases where longer-term dependencies must be modeled or a wide range of behaviors must be learned, we expect the Transformer or its lower-complexity alternative to perform better.

\begin{figure}[!htb]
	\centering
	\includegraphics[width=0.35\linewidth]{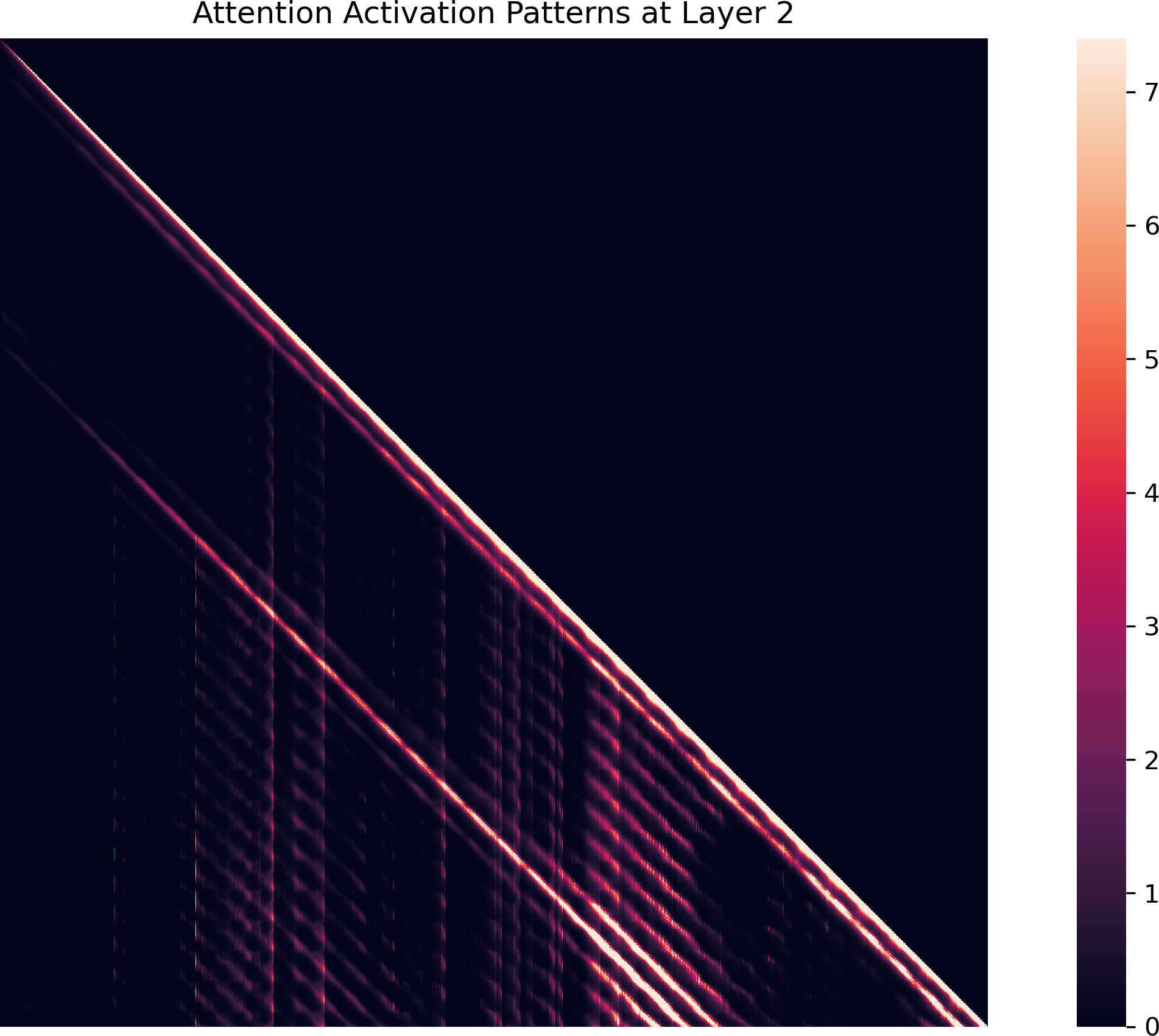}\hspace{2em}
	\includegraphics[width=0.35\linewidth]{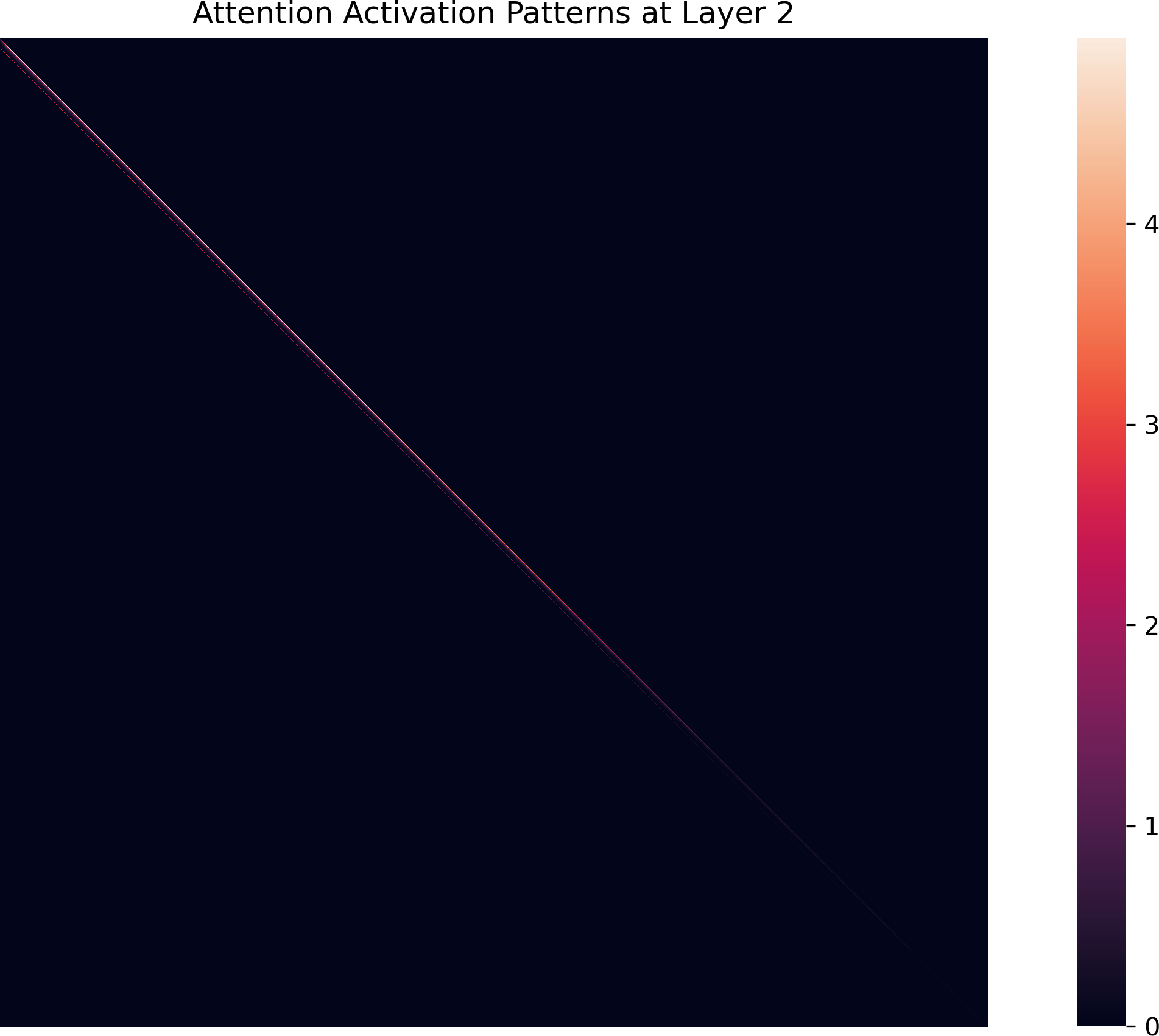}
	\caption{Attention activation patterns of the second layer of the Transformer (left) and Longformer (right) on the CPU validation set. The Transformer leverages positions that the Longformer masks. For the sake of readability, the attention activation patterns are compensated by multiplying each row with the number of unmasked positions.}
	\label{fig:cpu_attention_patterns}
\end{figure}

The Longformer performed significantly worse than the LSTM and Transformer on the CPU behavior. The attention patterns learned by the Transformer on this behavior depicted in Figure~\ref{fig:cpu_attention_patterns} reveal that the Transformer learns dependencies that span further than the window size of the Longformer while remaining in the range of what LSTMs can model~\citep{khandelwal2018sharp,dai2019transformer}. The Longformer has difficulty learning these dependencies with global tokens, which may be addressed by increasing their number or the window size.

\begin{figure}[!htb]
	\centering
	\includegraphics[width=\linewidth]{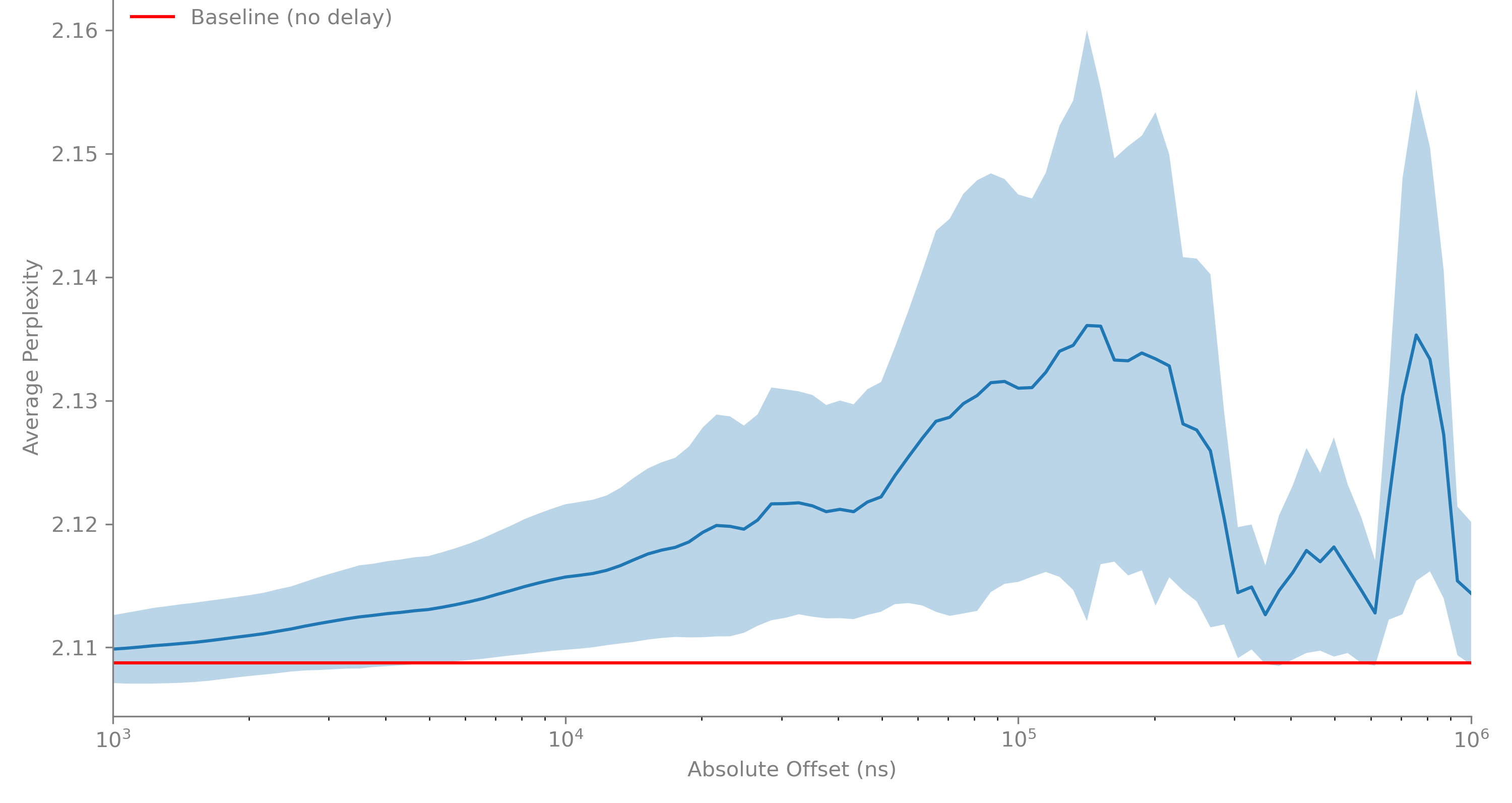}
	\caption{The average perplexity of an in-distribution request as 100 delays are introduced at 100 random positions each. The shade indicates the standard deviation, and the red horizontal line indicates the perplexity of the request without delays.}
	\label{fig:perplexity_delays}
\end{figure}

Delays from 1 microsecond ($\mu\text{s}$) to 1 millisecond (ms) were introduced randomly into an in-distribution sample to assess if the methodology can detect small latencies. The experiment involved first drawing a sample with $N$ system calls from the in-distribution validation set and evaluating its perplexity as the baseline. Then, the sample is duplicated $d \times p$ times, where $d$ is the number of delays and $p$ is the number of positions considered, and each of the $d$ delays is added to the duration of the system calls corresponding to each of the $p$ positions, thereby allowing to compute of the average perplexity for a given delay. Figure~\ref{fig:perplexity_delays} depicts the average perplexity as a function of the delay, which is always higher than the baseline and increases with the delay. Thus, the proposed methodology can detect small latencies, and its effectiveness increases as the delays increase. Additional figures for other in-distribution samples and language models are available on GitHub.

\section{Threats to Validity}
\label{sec:threats_to_validity}

This section acknowledges the potential threats to internal and external validity. Internal validity relates to the soundness of the evaluation methodology and its ability to draw conclusions from the results, while external validity relates to the ability to generalize the approach to other use cases.

\subsection{Threats to Internal Validity}

Two threats arise from evaluating a novelty detection methodology on datasets collected for that purpose. First, in-distribution and out-of-distribution sets may have unforeseen differences that facilitated the detection of novel behaviors, thus overestimating the methodology's performance. Second, in-distribution validation and test sets may contain behaviors not present in the training set, which can underestimate the methodology's performance. These threats are due to the high cost of manually investigating the dataset. They have been mitigated by collecting the datasets in a controlled environment with dedicated machines and comparing them with simple metrics, as shown in Table~\ref{tab:dataset}. The datasets and scripts have been publicly released for further investigation.

Another threat arises from the balanced nature of the validation and test sets, as novelties are rare in practice, and the number of positive samples is expected to be smaller than that of negative samples. Since the ratio of novelties depends on the use case, we leave it to researchers to sample positive instances based on their use case, and the harmonic mean of the precision and recall is reported.

The last threat arises from manually tuning the networks instead of conducting a grid or random search~\citep{bergstra2012random}, as it may lead to suboptimal hyperparameters. This approach is chosen to reduce the computational cost, thereby increasing reproducibility and minimizing the environmental impact. Despite the limited manual tuning, all the evaluated neural networks performed exceptionally well. As a mitigation, the code has been made publicly available and easily reproducible for researchers and practitioners to explore alternative architectures and hyperparameters.

\subsection{Threats to External Validity}

The leading threat arises from the low diversity of the datasets that may not represent real-world use cases, thus overestimating the generalization of the methodology. This is due to the lack of modern and massive public kernel trace datasets. Nonetheless, LSTMs and Transformers achieved comparable language model accuracy on a dataset similar to ours and a real-world dataset collected by Ciena~\citep{fournier2021on}, indicating that the collected dataset is representative of some real-world use cases. However, these two datasets contain a single behavior and are thus unsuitable for evaluating our novelty detection methodology. As a mitigation, the use cases were designed to be realistic and of genuine interest. Additionally, the trained models have been made public for researchers and practitioners to evaluate on their private datasets.

\section{Discussion}
\label{sec:discussion}

This section briefly discusses the strengths and acknowledges the limitations of the proposed methodology.

\subsection{Strengths and Benefits}

First, the approach is data-agnostic and novelty-agnostic. Indeed, language models are probability distributions over sequences of tokens, and while this work focuses on system calls as they reveal the system behavior without requiring manual instrumentation, tokens need not be limited to them. Furthermore, the detection only depends on the perplexity of the sequence under the language model; thus, any divergence from previously observed behavior will likely increase the perplexity, including hardware upgrades or failures, software updates, new users, rare queries, misconfigurations, latency, intrusions, and bugs.

Second, the approach does not heavily depend on an expert after the data collection to label the data or extract high-level features, which are often error-prone and suboptimal. While neural network hyperparameters must be tuned to achieve the best performance, techniques such as random search~\citep{bergstra2012random} have been developed to automate this process.

Finally, the approach is suitable for detecting novel behaviors in real-time as all three networks process a batch of 16 sequences in under 100ms on a single V100 GPU.

\subsection{Limitations and Shortcomings}

Neural networks are often considered to be black boxes, and their interpretability remains an active research topic. While the proposed approach cannot justify the detection, the predicted conditional probability of individual system calls may indicate the location of the root cause of the novelty.

Large language models may occasionally leak exact training samples~\citep{carlini2020extracting}. This potential privacy issue is not severe as one would need access to the model, only sequences of system call names without their arguments can be generated, and there is no indication of which of them are from the training data.

The Transformer is known to be sensitive to false negatives that are samples containing repeated strings~\citep{holtzman2019the}. A well-crafted attack that repeats a sequence of probable events may avoid detection. However, such attacks could be easily identified using simple metrics like the number of system calls or the request duration.

Finally, neural networks are computationally expensive and thus produce a large amount of carbon dioxide. \citet{strubell2020energy} estimated that training a Transformer with neural architecture search emits up to 284,000 kg of CO2. In comparison, the average American emits 16,400 kg annually, and the average car emits about 57,200 kg during its lifetime (fuel included). 

\section{Conclusion}
\label{sec:conclusion}

This paper introduces a new open-source dataset of kernel traces comprising over 2 million web requests with seven distinct behaviors and a new approach for detecting novelties based on language models. Surprisingly, the inductive bias of the LSTM toward local dependencies enabled more accurate novelty detection on 3 out of 6 behaviors compared to the more flexible Transformer and Longformer.

An essential direction is to continuously update the language models as the behavior of computer systems constantly evolves. Previous research has shown that language models can effectively learn from a small number of samples when trained at a sufficient scale~\citet{tsimpoukelli2021multimodal}. However, scaling neural networks increases the computation and memory required, preventing the detection of novelty in real-time. One promising avenue to scale the networks without increasing the amount of computation is a mixture of experts~\citep{jacobs1991adaptive}, which involves training multiple networks called experts and a router that forwards the input to a fixed number of relevant experts.

\section*{Acknowledgments}

We gratefully acknowledge the Natural Sciences and Engineering Research Council of Canada (NSERC), Prompt, Ericsson, Ciena, AMD, and EfficiOS for funding this project.



\end{document}